\documentclass[12pt]{article}

\usepackage{xspace}
\usepackage{latexsym}

\newcommand{\InferCred} 
 {\mbox{${\Infer}_{_{\vspace*{6mm}{\hspace*{-9pt}cred}}}$}}

\newcommand{\InferSL} 
  {\mbox{${\Infer}_{_{\vspace*{6mm}{\hspace*{-9pt}SL}}}$}}

\newcommand{\InferMTDR} 
  {\mbox{${\Infer}_{_{\vspace*{6mm}{\hspace*{-9pt}\textsl{MTDR}}}}$}}

\newcommand{\InferDelp} 
  {\mbox{${\Infer}_{_{\vspace*{6mm}{\hspace*{-9pt}\textsl{LP}}}}$}}

\newcommand{\InferNeg} 
  {\mbox{${\Infer}_{_{\vspace*{6mm}{\hspace*{-9pt}\textsl{neg}}}}$}}

\newcommand{\InferNot} 
  {\mbox{${\Infer}_{_{\vspace*{6mm}{\hspace*{-9pt}\textsl{not}}}}$}}

\newcommand{\InferNlp} 
  {\mbox{${\Infer}_{_{\vspace*{6mm}{\hspace*{-9pt}\textsl{NLP}}}}$}}

\newcommand{\qed}{\rule{2mm}{2mm}}

\newcommand{\comentario}[1]{}
\newcommand{\skippar}[1]{}

\newlength{\EA}  
\newlength{\MEA} 

\newlength{\premisas}
\newlength{\consec}
\newlength{\rotulo}
\newlength{\largoregla}

\newtheorem{Definicion}{{\sc Definici\'on}}[section]

\newtheorem{DefinitionText}{{\sc Definition}}[section]

\newtheorem{Ejemplo}{{\sc Ejemplo}}[section]

\newtheorem{ExampleText}{{\sc Example}}[section]

\newtheorem{Proposicion}{{\sc Proposici\'on}}[section]

\newtheorem{PropositionText}{{\sc Proposition}}[section]

\newtheorem{Corolario}{{\sc Corolario}}[section]

\newtheorem{CorollaryText}{{\sc Corollary}}[section]

\newtheorem{Lemax}{{\sc Lema}}[section]

\newtheorem{LemmaText}{{\sc Lemma}}[section]

\newtheorem{Teox}{{\sc Teorema}}[section]

\newtheorem{TheoremText}{{\sc Theorem}}

\newtheorem{Notax}{{\bf Observaci\'on}}[section]

\newtheorem{obsx}{{\bf Observaci\'on}}[section]

\newtheorem{Algox}{{\sc Algoritmo}}[section]

\newlength{\blk}
\newcommand{\Infer}{\mbox{$\mid\hspace{-2pt}\sim\ $}}

\newcommand{\ArgumGeneral}[1]{\mbox{$\rangle {\cal A}, h \langle$}}

\newcounter{nro-regla}
\newenvironment{regla-infer}{\begin{center}}{\end{center}
                         \addtocounter{nro-regla}{1}}

\newcommand{\personal}[1]{}

\newcommand{\comentado}[1]{}

\begin{document}

\begin{center}
     {\Large\bf      Integrating Defeasible Argumentation \\
                      and Machine Learning Techniques  \\
                                    \vspace{1mm}
                         {\large (Preliminary Report)} }
\end{center}

\begin{center}
    { \large Sergio Alejandro G\'{o}mez \hspace{1.5cm}    Carlos Iv\'{a}n
    Ches\~{n}evar }
\end{center}

\begin{center}
\begin{tabular}{c}
{ Laboratorio de Investigaci\'on y Desarrollo en Inteligencia
Artificial}
\\
{Departamento de Ciencias e Ingenier\'{\i}a de la
Computaci\'on}  \\ {\sc Universidad Nacional del Sur}
\\ {Av.\ Alem 1253 -- B8000CPB Bah\'{\i}a Blanca -- {\sc Rep\'ublica
Argentina}} \\ {\sc Tel/Fax:} (+54) (291) 459 5135/5136 -- {\sc
Email:} {\tt \{sag, cic\}@cs.uns.edu.ar} \\
\end{tabular}

 \vspace{2mm}
\begin{footnotesize}
\end{footnotesize}
\end{center}

\nocite{ChesnePHD2001}
\nocite{Simari92}
\nocite{GarciaPHD2000}
\nocite{Survey2000}
\nocite{Tou2001}


\begin{abstract}
\noindent The field of machine learning (ML) is concerned with the
question of how to construct algorithms that automatically improve
with experience. In recent years many successful ML applications
have been developed, such as datamining programs,
information-filtering systems, etc. Although ML algorithms allow
the detection and extraction of interesting patterns of data for
several kinds of problems, most of these algorithms are based on
\emph{quantitative} reasoning, as they rely on training data in
order to infer so-called target functions.

In the last years \emph{defeasible argumentation} has proven to be
a sound setting to formalize common-sense \emph{qualitative}
reasoning. This approach can be combined with other inference
techniques, such as those provided by machine learning theory.

In this paper we outline different alternatives for combining defeasible
argumentation and machine learning techniques. We suggest how different
aspects of a generic argument-based framework can be integrated with
other ML-based approaches.
\end{abstract}

\section{Introduction and motivations}
\label{sec:intro}

The ability to generate and collect data has increased exponentially in
the last years. Automatizing transactional databases has resulted in
an explosive growth of information, motivating the evolution of
several \emph{datamining} techniques.
Simply stated, datamining refers to \emph{extracting or mining knowledge
from large amounts of data}. This knowledge constitutes non-trivial
and potentially useful information that can be obtained in many cases
by applying \emph{machine learning} (ML) techniques.

The field of machine learning is concerned with the question of how
to construct algorithms that automatically improve with experience~\cite{mitchell}.
In recent years many successful ML applications have been developed,
such as datamining programs, information-filtering systems, etc.
Although ML algorithms allow the detection and extraction of
interesting patterns of data for several kinds of problems,
most of these algorithms provide an output based
on \emph{quantitative} evidence (i.e. training data), whereas
the inference process which  led to this output is commonly unknown
(i.e. `black-box' metaphor).

In the last years \emph{defeasible
argumentation}~\cite{Survey2000,PraVre99,Simari92} has proven to be
a sound setting to formalize common-sense \emph{qualitative}
reasoning. This approach can be combined with other inference
techniques, such as those provided by machine learning theory.

In this preliminary report we explore different alternatives for
developing applications which combine defeasible argumentation and
machine learning techniques. We
suggest how different aspects of a generic argument-based
framework can be integrated with other ML-based
approaches. The paper is structured as follows. First, we briefly
introduce the components of most argument-based framework and then
outline possible directions for the integration of ML techniques
and defeasible argumentation. Next, we describe a particular
setting suitable for the application of such approach, namely
text mining problems.
Finally, we discuss some promising research lines that are
currently being pursued.

\section{Integrating ML and argument-based frameworks}
\label{sec:framework}

Argument-based frameworks~\cite{Simari92,Survey2000,PraVre99}
provide a sound formalization of defeasible reasoning, and have
found a wide acceptance in many areas such as development of legal
reasoning applications, multiagent systems, etc. As pointed out
in~\cite{Carbogim00}, most argument-based frameworks share a number
of common notions, namely:

\begin{enumerate}
\item
  \textbf{Knowledge Base formalized in an underlying logical language}:
  Most argument-based frameworks involve
  a knowledge base $K=(\Pi,\Delta)$ which provides \emph{background knowledge}
  for an intelligent agent formalized in a first-order logical language $L$.
  This background knowledge typically involves a set $\Pi$ of \emph{strict rules}
  and \emph{facts}  and a set $\Delta$ of \emph{defeasible rules}.

\item
   \textbf{Argument}:
   An  \emph{argument} is a defeasible proof obtained from the knowledge
   base $K$ by applying suitable (defeasible) inference rules
   associated with the underlying logical language $L$.

\item \textbf{Dialectical reasoning}:
   Given two arguments $A$ and $B$, conflict (or attack) among arguments
   arises whenever  $A$ and $B$ cannot be simultaneously accepted (typically
   because of some kind of logical contradiction). Many argument systems provide a
   preference criterion which defines a partial order among arguments,
   allowing to determine whether $A$ should be preferred over $B$.
   This defines a \emph{defeat} relationship.
   Given the set $Args$ of arguments obtained from a knowledge base $K$,
   it holds that $attacks \subseteq Args \times Args$, and  $defeats \subseteq attacks$.
   In order to determine whether a given argument $A$ is ultimately
   undefeated (or \emph{warranted}), a dialectial process is recursively
   carried out, where defeaters for $A$, defeaters for these defeaters, and so on,
   are taken into account.
 \end{enumerate}

Next we will outline different approaches that we are currently considering
to model the above issues in the context of ML techniques.

\begin{itemize}

\item
   \textbf{Building a defeasible knowledge base from training data}:


Recently there have been some approaches to obtaining defeasible
rules from training data, particularly in the context of Inductive
Logic Programming (ILP). In~\cite{Inoue}, the authors present a
method to generate non-monotonic rules with exceptions from
positive/negative examples and background knowledge in Inductive
Logic Programming. The form of the programs to be learnt is the
one of \emph{extended logic programs}, which incorporates both
strict negation and negation as failure that can be effectively
used in the presence of incomplete information. Default rules are
generated as specializations of general rules that cover positive
examples, whereas exceptions to general rules are identified from
negative examples and are then generalized to rules for
cancellation of defaults. The resulting learning system LELP
allows also to learn hierarchical defaults by recursively calling
an exception identification algorithm.


In~\cite{Guido}, the authors investigate
the feasibility of
Knowledge Discovery from Data-bases (KDD) in order to facilitate
the discovery of defeasible rules for legal decision making. In
particular they argue in favour of Defeasible Logic as an
appropriate formal system in which the extracted principles should
be encoded.

In the context of obtaining defeasible rules  by means of induction-based techniques,
the work of Peter Flach~\cite{Flach} provides an interesting survey of
several approaches to computational induction and provides a descriptive
theory of induction as a logical framework system.
Several rule systems for conjectural reasoning are axiomatized
and semantically characterized.

\item
\textbf{Building arguments}:
\newcommand{\argument}[2]{\langle #1, #2 \rangle}

Finding hypotheses which explain unseen instances is
central to most ML algorithms. Interestingly, in argument-based
frameworks an argument is also understood as
providing ``a hypothesis supporting a given conclusion''~\cite{Simari92}.
We contend that argument-based reasoning and
many ML techniques share this common notion. Many theoretical
results from argument theory could therefore be applied in a ML context.

\emph{Analytical learning} methods~\cite{mitchell}
(like explanation-based learning) offer the advantage of
generalizing more accurately from less data by using prior
knowledge to guide learning.  However, they can be misled when
given incorrect or insufficient prior knowledge.  On the other
hand, \emph{inductive learning} methods (like neural nets, decision tree
learning, inductive logic programming) offer the advantage that
they require no explicit prior knowledge and learn regularities
based solely on the training data. However, they can fail when
given insufficient training data, and can be misled by the
implicit inductive bias they must adopt in order to generalize
beyond the observed data.
A combined analytical and inductive ML method that overcomes the pitfalls
associated with each separate approach (yet conserving their
individual advantages) should be as follows: given a set of
training examples $D$ of some target function $f$ (possibly
containing errors), a domain theory $B$ (possibly containing
errors), and a space of candidate hypothesis $H$, determines which
hypothesis $h$ fits best the training examples and domain theory.

We think that defeasible argumentation can be used to improve ML
approaches as the one described above  as it provides a sound
formalization for both expressing and reasoning with uncertain and
incomplete information. From the set of examples $D$ and theory
$B$, a knowledge base $(\Pi, \Delta)$ can be induced as explained
above. The hypothesis $h$ can be expressed as an argument structure
$\argument{A}{h}$ where $A$ stands for both the relevant part of
the background theory and relevant features of $D$.

\item
\textbf{Dialectical reasoning and ML}:

The process of defeasible argumentation always involves the
analysis of conflicting arguments in a dialectical setting. As
explained before, such setting relies on a \emph{defeat}
relationship for comparing conflicting arguments. Determining
whether an argument is ultimately accepted requires a recursive
analysis in which defeaters, defeaters for these defeaters, and so
on, are taken into account~\cite{Survey2000}.

In many formalizations of argumentation the notion of defeat is
considered as an abstract relationship (i.e. a partial order $\preceq$
among  arguments). Some approaches (e.g.~\cite{Simari92})
rely on the notion of \emph{specificity} for defining defeat.
In this context, ML techniques provide sound alternatives for
considering \emph{numeric attributes} or \emph{probabilistic values}
for deciding between conflicting hypotheses~\cite{mitchell}.
We think that such approaches could provide the basis for defining
new comparison criteria in the context of argument-based frameworks.
In the same line of reasoning, contrasting conflicting hypotheses is
a common situation in many ML algorithms.
In this setting defeasible argumentation provides a useful theoretical background
for contrasting such hypotheses,
making easier to identify \emph{fallacious} patterns of reasoning
which might lead to incorrect results.


\end{itemize}

\section{Text mining: a case study}
\label{sec:textmining}


Datamining is the process of discovering interesting patterns
from large amounts of data \cite{han}. In \emph{textual datamining} the needs
of the user may vary from looking for a specific piece of text
to getting familiarized with a subject area \cite{honkela}.
Basic approaches for information retrieval and
data mining involve the following strategies:
\begin{itemize}
     \item
     \emph{Searching using keywords:} This is done by
    automatically indexing the documents by frequency of term
    appearance.
    This approach is used by web crawlers \cite{brin}
     and traditional information retrieval systems \cite{frakes}.
    \item
     \emph{Exploration of the document collection supported by
    organizing the documents in some manner:} For each document,
    an internal representation is obtained
    and then the documents are
    arranged into clusters by some similarity measure \cite{rasmussen}.
    This process can be done by using ML
    techniques such as neural networks \cite{gomez1, gomez2, honkela,
    kohonen} or bayesian classifiers \cite{billsus, pazzani}.
     \item
     \emph{Filtering:}  It refers to discarding uninteresting
    documents from an incoming document flow \cite{mostafa}.
 \end{itemize}

Each approach has its pros and cons. Searching using keywords is
easy to implement but can lead to retrieve unrelated documents or,
even worse, to not discover related documents (this can be
measured using some metric such as precision and recall
\cite{frakes}). Clustering of documents can be efficiently
implemented by neural nets but may require retraining in the
presence of new examples; besides, in this case the model is not
understandable by a human programmer because it is compiled as a
set of numerical weights.  Filtering is difficult because of the
dynamic nature of user interests and document flow \cite{gomez1}.

Some recent approaches propose using argumentation to analyze
structured text in the form of XML documents~\cite{besnard-hunter,
hunter1}. As explained in the previous section, we feel that the
integration of defeasible argumentation and ML can tackle many of
the problems described above, thus enhancing existing algorithms
for text mining.

\section{Conclusions and future work}
\label{sec:conclu}

The success of argumentation-based approaches is partly due to the
sound setting it provides for \emph{qualitative reasoning}.
Although numeric attributes offer an useful source of information
for \emph{quantitative reasoning} in several knowledge domains,
they have been mostly neglected  in the defeasible argumentation
community. This is maybe due to the historical origins of
argumentative reasoning, which were more related to legal
(qualitative) reasoning rather than to number-based attributes as
those used in rule-based production systems.

We think that integrating ML techniques with argumentation
frameworks would be highly desirable, as it would provide a
combination of both analytical and inductive ML methods capable of
tackling the pitfalls of each separately yet conserving their
advantages, making them more attractive and suitable for other
research and application areas. Part of our current research work
is focused on these aspects.


\begin{footnotesize}
\bibliographystyle{giia}
\newcommand{\etalchar}[1]{$^{#1}$}

\end{footnotesize}

\end{document}